\documentclass[runningheads]{llncs}

\usepackage[T1]{fontenc}
\usepackage{graphicx}
\usepackage{amsmath,amsfonts}
\usepackage{booktabs}
\usepackage{tabularx}
\usepackage{adjustbox}
\usepackage{float}
\usepackage{url}
\usepackage{microtype}
\usepackage{amsmath}
\usepackage{amssymb}
\usepackage{cite}
\usepackage{hyperref}
\usepackage{cleveref}

\begin{document}
\raggedbottom

\title{Large-Scale Pretraining for Improving Deep Learning-Based Geometric Distortion Correction of Diffusion-Weighted Imaging}

\titlerunning{Large-scale Pretraining for DWI Distortion Correction}

\author{
Saroj Khanal\inst{1} \and
Yashawant Kumar Yadav\inst{2} \and
Kritam Bhattarai\inst{3} \and
Jeevan Neupane\inst{4} \and
Shristi Subedi\inst{5} \and
Saship Gwachha\inst{5} \and
Manish Kumar Tiwari\inst{6} \and
Dong Zhang\inst{7, 8} \and
Confidence Raymond\inst{7, 9, 10} \and
Aondona Moses Iorumbur\inst{7, 11} \and
Udunna Anazodo\inst{7, 9, 10} \and
Surendra Maharjan\inst{12} \and
Bishesh Khanal\inst{1} \and
Mahesh Shakya\inst{1} \and
Pralhad Kumar Shrestha\inst{1}\orcidID{0000-0001-5831-4086}
}

\authorrunning{S. Khanal et al.}
\institute{
Nepal Applied Mathematics and Informatics Institute for Research (NAAMII), Lalitpur, Nepal \and
University of Exeter, Exeter, Devon, UK \and
Institute of Engineering, Purwanchal Campus, Dharan, Nepal \and
Institute of Engineering, Pulchowk Campus, Lalitpur, Nepal \and
Gandaki Medical College Teaching Hospital and Research Center, Pokhara, Nepal \and
Madan Bhandari University of Science and Technology, Chitlang, Nepal \and
Medical Artificial Intelligence Laboratory, Lagos, Nigeria \and
Department of Electrical and Computer Engineering, University of British Columbia, Vancouver, Canada \and
Montreal Neurological Institute, McGill University, Montreal, Canada \and
Department of Biomedical Engineering, McGill University, Montreal, Canada \and
Department of Medical Physics, College of Medicine, University of Lagos, Nigeria \and
Brain Health Imaging Institute, Department of Radiology, Weill Cornell Medicine, Weill Cornell Medicine, New York, United States \and
\email{\{saroj.khanal, bishesh.khanal, mahesh.shakya\}@naamii.org.np},
\email{y.yadav@exeter.ac.uk},
\email{\{pralhad.shrestha05, kritambhattarai4, jeevan.neupane003, srii.sbd, gwachhasaship, mosesiorumbur, suren634634\}@gmail.com},
\email{manish.kumar.tiwari@mbust.edu.np, udunna.anazodo@mcgill.ca, confidence.raymond@mail.mcgill.ca, donzhang@ece.ubc.ca}
}

\maketitle

\begin{abstract}
\begin{sloppypar}

Diffusion-weighted imaging (DWI) is widely used in clinical settings but remains vulnerable to geometric distortion. 
Conventional correction methods often require additional acquisitions or vendor-specific solutions, limiting their feasibility in high-throughput, resource-constrained settings.
This study investigates whether large-scale pretraining strategies can improve deep learning-based distortion correction for single-phase-encoding DWI. 
We formulate the task as image reconstruction, and compare a non-pretrained baseline against a self-supervised and a generative pretrained model, evaluated using both quantitative image-similarity metrics and qualitative expert assessment. 
The best-performing model was further tested for transferability on data collected in an LMIC setting with acquisition shift.
Pretrained models outperformed the non-pretrained baseline, with cWDM achieving the strongest results across both quantitative and qualitative evaluation. 
However, application to LMIC data revealed transferability challenges, including contrast alteration and over-reliance on T1-weighted anatomical structure. Registering images to a common standard space improved predictions, suggesting that harmonized preprocessing may enhance cross-domain deployment.

\end{sloppypar}

\keywords{Diffusion-weighted imaging \and SS-EPI \and Geometric distortion correction \and Large-scale pretraining \and Conditional wavelet diffusion model}

\end{abstract}

\section{Background}

In routine clinical settings, Diffusion-weighted imaging (DWI) is the gold standard for identifying acute ischemic stroke, often within minutes of symptom onset and before conventional MRI becomes clearly abnormal \cite{le2024brownian}. 
It also helps detect tumors, distinguish malignant from benign lesions \cite{karmakar2023utility}, and monitor treatment response across multiple organ systems in cancer imaging \cite{messina2020diffusion}.

Single shot echo planar imaging (SS-EPI) is the most widely used acquisition technique and serves as a standard pulse sequence for DWI due its high temporal efficiency. 
However, the technique is inherently susceptible to significant  geometric distortion and blurring caused by magnetic field inhomogeneities \cite{ye2021simultaneous, zhang2023diagnostic}. 
Multi-shot MRI sequences, such as RESOLVE by Siemens \cite{zhong2014resolve}, are specifically designed to reduce the geometric distortions and blurring artifacts commonly seen in ss-EPI acquisitions. 
However, these techniques are often expensive and may require proprietary licensing, making them less suitable for Low- and Middle-Income Countries (LMIC) settings.

Although traditional imaging methods, such as TOPUP \cite{andersson2003correct}, mitigate these distortions and artifacts, they often require additional reference scan from opposite phase encoding direction or lengthy acquisition times \cite{aamir2022accelerated, chen2021qsm}. 
These requirements can reduce patient throughput, make them unsuitable for retrospectively collected scans, and limit their practicality in resource-limited clinical settings.

Towards addressing this gap, this study explores deep learning-based distortion correction of DWI images using large-scale pretraining strategies followed by model performance evaluation through qualitative assessment by radiographers. Furthermore, to simulate a LMIC clinical setting with limited access to advanced acquisition techniques, constraints on lengthy acquisition protocols, and insufficient training data, we evaluate the model’s transferability using LMIC data collected with different acquisition protocols and demographic characteristics.

Our Contribution is three-fold:
\begin{enumerate}
\item We leverage large-scale, open-weight MRI foundation models, specifically incorporating diffusion-based conditional generative pretraining alongside self-supervised pretraining, to systematically evaluate their advantages against baseline, non-pretrained AI models.

\item We perform qualitative evaluations by clinical experts to complementarily measure downstream relevant semantics of the image along with image pixel-based metrics.

\item We explore transferability of the best-performing model by evaluating them on a locally-collected DWI distortion dataset from real clinical setting and evaluate them via expert-based qualitative evaluation.

\end{enumerate}

\section{Previous Works}
Deep learning algorithms have emerged as promising alternative for DWI distortion correction from single-phase encoding acquisitions\cite{dargahi2025susceptibility,schilling2019synthesized}.
However, generalization to unseen scanners and acquisition contexts remain limited.
Additionally, multiple evidences has shown that large-scale pretraining improves performance on downstream tasks in medical imaging\cite{wang2023mis,mirugwe2025improving,wu2022skin}.
We extend this underexplored direction for DWI distortion correction.
In addition, DWI correction algorithms are often reported using Image-only metrics, such as mean squared error, which may fail to capture certain clinically-relevant semantics\cite{borasinski2022paired}.
We accordingly incorporate qualitative assessment to evaluate these models.

\section{Methods}

\subsubsection{Study Design}
\begin{figure}[!t]
\centering
\vspace{-0.4em}

\includegraphics[
    width=0.8\textwidth
]{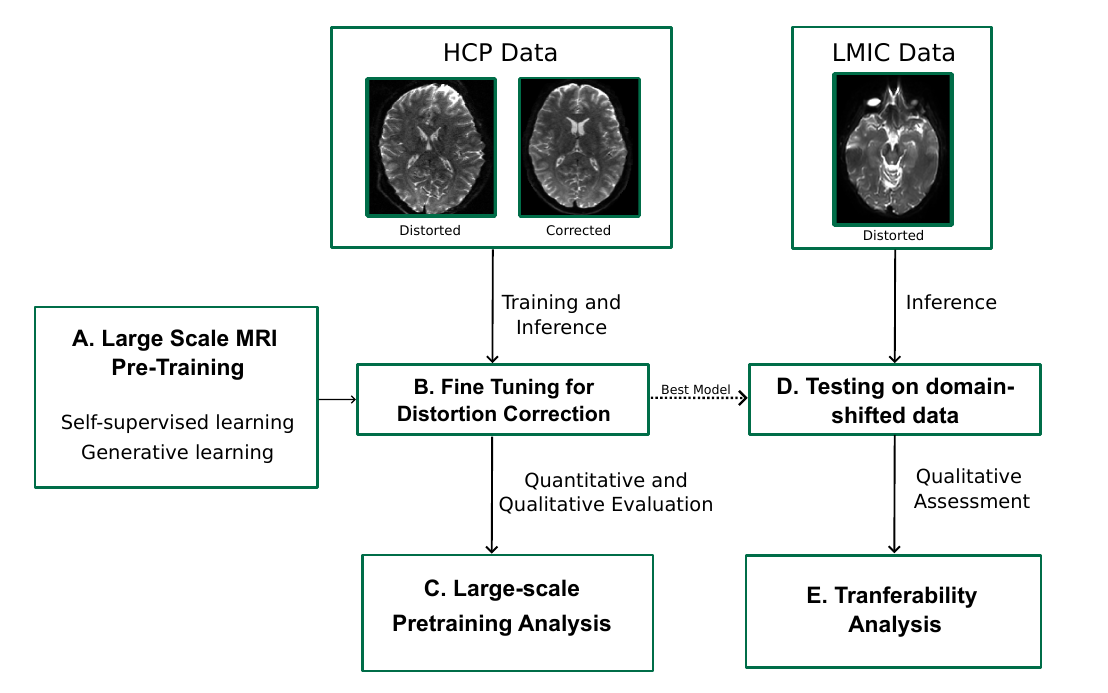}

\vspace{-0.6em}
\caption{Block diagram of the proposed methodology. A) Large-scale MRI pretrained models were obtained from publicly available repositories. B) The pretrained models were fine-tuned for DWI distortion correction. C) Pretraining effectiveness was assessed through quantitative and qualitative evaluation. D) The best-performing model was tested on domain-shifted LMIC clinical data. E) Model transferability was evaluated through qualitative expert assessment.}
\label{fig:methodology}
\vspace{-0.5em}
\end{figure}

We formulate the distortion correction task as supervised image-to-image reconstruction, learning $f_\theta : (X_{b0}, X_{T1}) \rightarrow Y_{b0}$, where $X_{b0},X_{T1},Y_{b0}\in \mathbb{R}^{H \times W \times D}$, denotes the distorted b0 image, T1-weighted reference image, and the ground-truth corrected b0 target images, respectively, over dataset $\mathcal{D} =
\left\{
(X_{b0}^{i}, X_{T1}^{i}, Y_{b0}^{i})
\right\}_{i=1}^{N}$. We train a non-pretrained baseline and fine-tune two models pretrained on large-scale MRI datasets under the same formulation, evaluating all three on the held-out test set via quantitative similarity metrics and radiographer assessment. The best-performing model is further applied to LMIC-acquired data with marked domain shift in acquisition and population, followed by expert qualitative evaluation (Fig.~\ref{fig:methodology}).

\subsubsection{Dataset and Preprocessing}

Two datasets are used in this study: the publicly available Human Connectome Project (HCP) dataset\cite{vanessen2013hcp} and a retrospectively collected Nepalese dataset representing a real-world clinical LMIC setting with single phase-encoding (PE) acquisition. 
These two datasets differ in acquisition protocol and population representation, with HCP data acquired along the left-right direction from a Western population and the Nepalese dataset acquired along anterio-posterior direction representative a region from the Global South.

The HCP dataset consists of paired T1-weighted and spin-echo EPI sequence(ss-EPI) DWI images from 1065 subjects. 
To correct for severe geometric distortions inherent to ss-EPI, HCP acquires data with opposite phase encoding directions (Right-Left and Left-Right) to compute field maps and perform unwarping.
After this postprocessing, a corrected DWI image is obtained acting as groundtruth.

On the other hand, the Nepalese dataset consists of paired T1-weighted and ss-EPI DWI Images without additional reverse-phase encoding, collected retrospectively from a real clinical workflow in LMIC setting.
This single-phase encoding means a corrected DWI image as groundtruth is not available.
Hence, we perform expert-based qualitative evaluation.

For training and evaluation, all image volumes were resampled to a uniform physical resolution of $1 \times 1 \times 1~\mathrm{mm}$ with bilinear interpolation and resized to $96 \times 96 \times 96$ voxels using the MONAI \cite{cardoso2022monai} framework. Image intensities are normalized using min--max scaling, which mapped the intensity range to $[-1, 1]$. FSL's Brain Extraction Tool (BET) \cite{smith2002fast} was used to generate brain masks, and the loss was calculated only within the brain region.

\subsubsection{Benchmarked Models}
\textit{Synb0-DISCO Baseline}
Synb0-DISCO \cite{schilling2019synthesized}, a supervised 3D U-Net for single phase-encoding distortion correction, serves as our state-of-the-art benchmark. We train it from scratch (no pretrained initialization) to isolate the contribution of large-scale MRI pretraining.

\noindent\textit{SwinBRAIN}
SwinBRAIN \cite{suo2025automatic} provides self-supervised Swin-UNETR weights pretrained on 75,861 multi-modal head MRI scans (T1, T2, FLAIR). We fine-tune these publicly available weights for b0 distortion correction, testing whether anatomical representations learned across MRI modalities transfer to this task.

\noindent\textit{Conditional Wavelet Diffusion Model (cWDM)}
cWDM \cite{friedrich2024cwdm} is a conditional generative model for cross-modality 3D MRI synthesis, performing wavelet-space image-to-image translation via a U-Net-like denoising backbone. We use weights pretrained on BraTS 2021 ($\sim$6,000 scans) for T2 synthesis from T1, contrast-enhanced T1, and FLAIR\cite{pfriedri_cwdm}, hypothesizing this paired-synthesis initialization benefits distortion correction. We adapt cWDM for deterministic reconstruction: distorted b0 and T1 images are wavelet-transformed, passed through the pretrained backbone, and inverse-transformed to yield the corrected b0 image.

\subsubsection{Training Strategy}

The dataset is split into train, validation, and test sets in a 72:13:15 ratio. The models were trained with a maximum training budget of 30 epochs on an NVIDIA GTX 1070Ti. All models converged with the following hyperparameters: batch size 1, Adam optimizer with learning rate $1 \times 10^{-4}$ and weight decay $1 \times 10^{-5}$, and mean square error loss with spatial masking to avoid the background region.

\subsubsection{Quantitative Evaluation}
Quantitative evaluation is performed on the held-out HCP test set by comparing each model-generated corrected b0 image with the corresponding HCP corrected b0 target. Five image-similarity metrics were calculated: mean squared error (MSE), peak signal-to-noise ratio (PSNR), structural similarity index measure (SSIM) \cite{wang2004image}, visual information fidelity (VIF) \cite{sheikh2006image}, and spectral angle mapper (SAM). These are reference-based visual similarity metrics commonly used in distortion correction literature. All metrics were calculated within the brain region using the BET-derived brain masks. Metrics were calculated for each test subject, and the Wilcoxon signed-rank test was used for pairwise comparison of model performance across all quantitative metrics.

\subsubsection{Qualitative Evaluation}

To assess correction fidelity beyond voxel-based metrics, we developed a clinician-centered evaluation protocol targeting anatomy preservation, texture integrity, and distortion correction, in absence of reference corrected images. The protocol was developed by consensus among three radiographers ($>$3 years experience) and a senior radiology associate.
Distortion correction was scored on a 5-point Likert scale (5 = no distortion, with preserved anatomical boundaries and no visible susceptibility-related artifacts; 4 = mild distortion, with minor geometric distortion or artifacts but preserved diagnostic quality; 3 = moderate distortion, with visible distortion in regions such as the frontal or temporal lobes, parietal regions, or brainstem, while major structures remain identifiable; 2 = severe distortion, with substantial geometric distortion, signal dropout, and loss of anatomical detail; and 1 = complete distortion, where the image is non-diagnostic.)
Spatial data integrity, evaluated against the T1-weighted image, was scored as excellent/good/fair/poor based on severity of data modification.
Excellent indicated no visible data modification, good indicated slight but acceptable artifacts, fair indicated moderate and distracting image modification, and poor indicated severe data corruption or non-diagnostic image quality.
Each model's test-set predictions were rated by one blinded radiographer (model identity and quantitative scores withheld); 50 samples were additionally rated by all three raters to assess inter-rater agreement via Gwet's AC2 \cite{walsh2014approaches}, chosen for robustness to the kappa paradox under skewed distributions. Spearman correlation was computed between quantitative and qualitative scores to assess their association.

\section{Result and Discussion}

\subsubsection{Results on HCP Data}

As shown in Table~\ref{tab:hcp_quantitative_performance}, both pretrained-then-fine-tuned models significantly outperformed the non-pretrained baseline across all metrics (p<0.01), with the generative model (cWDM) performing best despite its smaller pretraining dataset - suggesting pretraining on a paired MRI synthesis task transfers more effectively. 
An ablation training cWDM from scratch confirmed this: pretrained cWDM outperformed its non-pretrained counterpart across all metrics (p < 0.01) - MSE 0.008 vs. 0.011, PSNR 34.009 vs. 32.726, SSIM 0.920 vs. 0.899, and VIF 0.504 vs. 0.460 - underscoring the benefit of pretraining.
In terms of computational cost, cWDM required 10.091 ± 0.645 s per volume on CPU and 0.223 ± 0.002 s on a T4 GPU, with a peak GPU memory footprint of 673.89 MB.

\begin{table}
\centering
\caption{Quantitative performance on the HCP test set.}
\label{tab:hcp_quantitative_performance}
\begin{adjustbox}{width=\textwidth}
\begin{tabular}{lcccc}
\toprule
Model Name & MSE $\downarrow$ & PSNR $\uparrow$ & SSIM $\uparrow$ & VIF $\uparrow$ \\
\midrule
Baseline Non-Pretrained Synb0-DISCO & 0.017 $\pm$ 0.005 & 30.791 $\pm$ 1.232 & 0.791 $\pm$ 0.026 & 0.480 $\pm$ 0.030 \\
Self-Supervised Pretrained SwinBRAIN & 0.015 $\pm$ 0.005 & 31.293 $\pm$ 1.473 & 0.864 $\pm$ 0.029 & 0.406 $\pm$ 0.028\\
Generative Pretrained cWDM & \textbf{0.008 $\pm$ 0.003} & \textbf{34.009 $\pm$ 1.538} & \textbf{0.920 $\pm$ 0.021} & \textbf{0.504 $\pm$ 0.025} \\
\bottomrule
\end{tabular}
\end{adjustbox}
\end{table}
\begin{figure}
    \centering
    \includegraphics[width=0.75\textwidth]{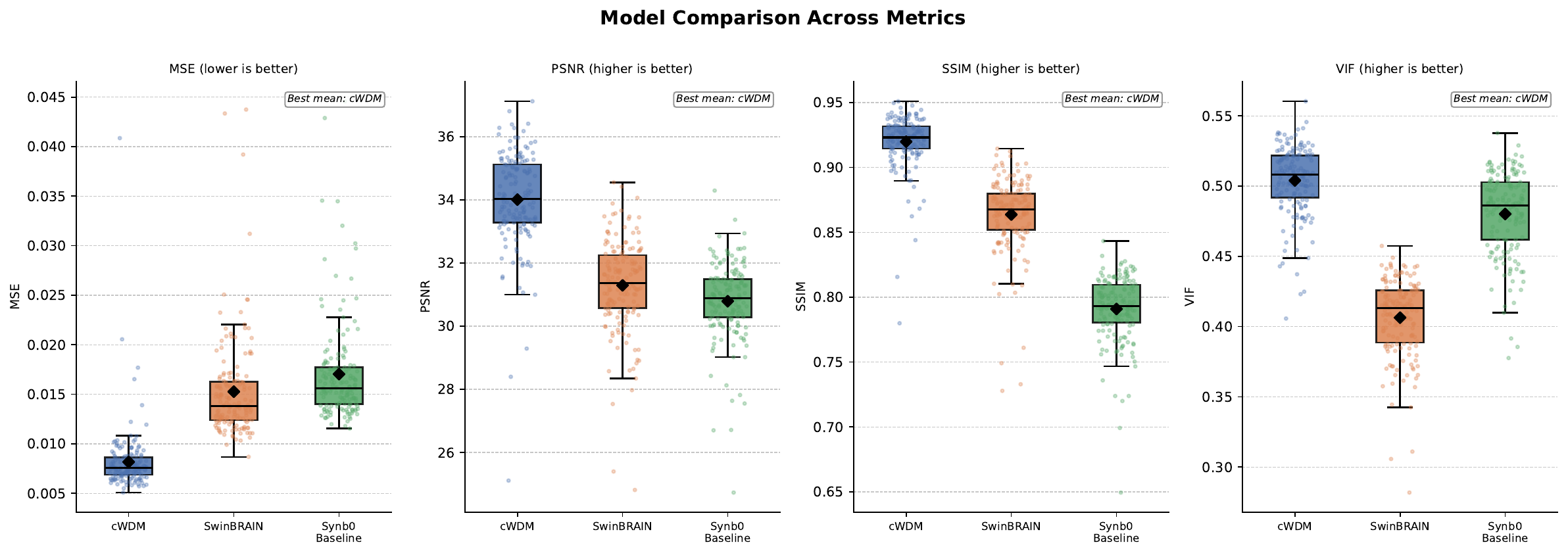}
    \caption{Box plots representation of quantitative metrics across three models on HCP test set.}
    \label{fig:figure2}
\end{figure}

\begin{figure}[!htb]
    \centering
    \begin{minipage}[t]{0.48\textwidth}
        \centering
        \includegraphics[width=\linewidth]{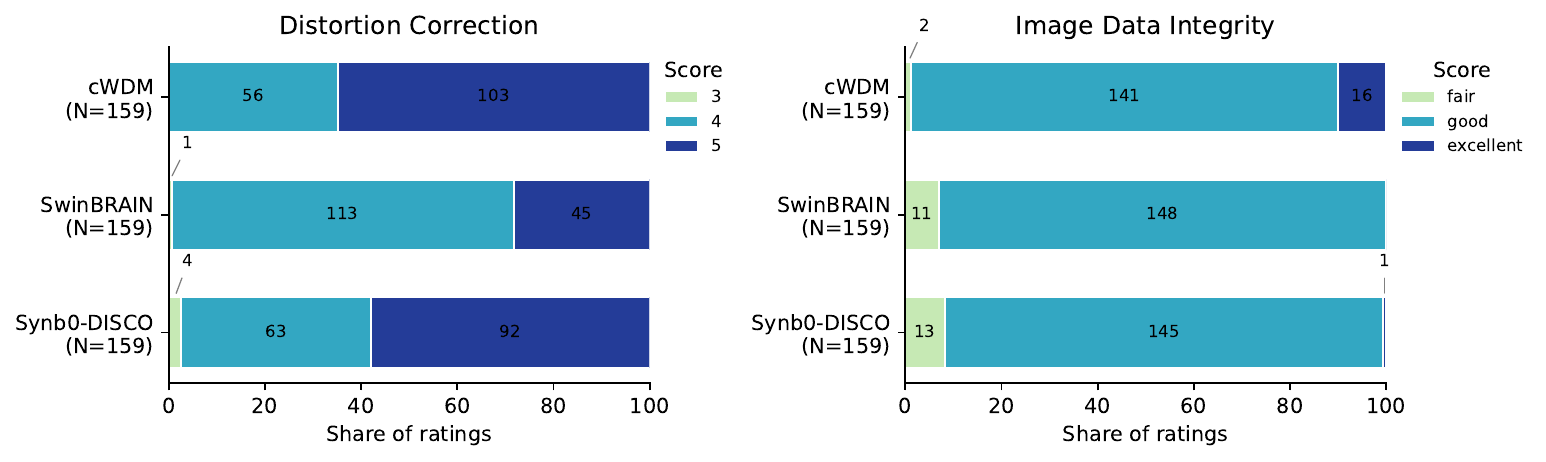}
        \vspace{0.5em}
        \textbf{(a)} Radiographer ratings
        \label{fig:radiographer_ratings}
    \end{minipage}
    \hfill
    \begin{minipage}[t]{0.48\textwidth}
        \centering
        \includegraphics[width=\linewidth]{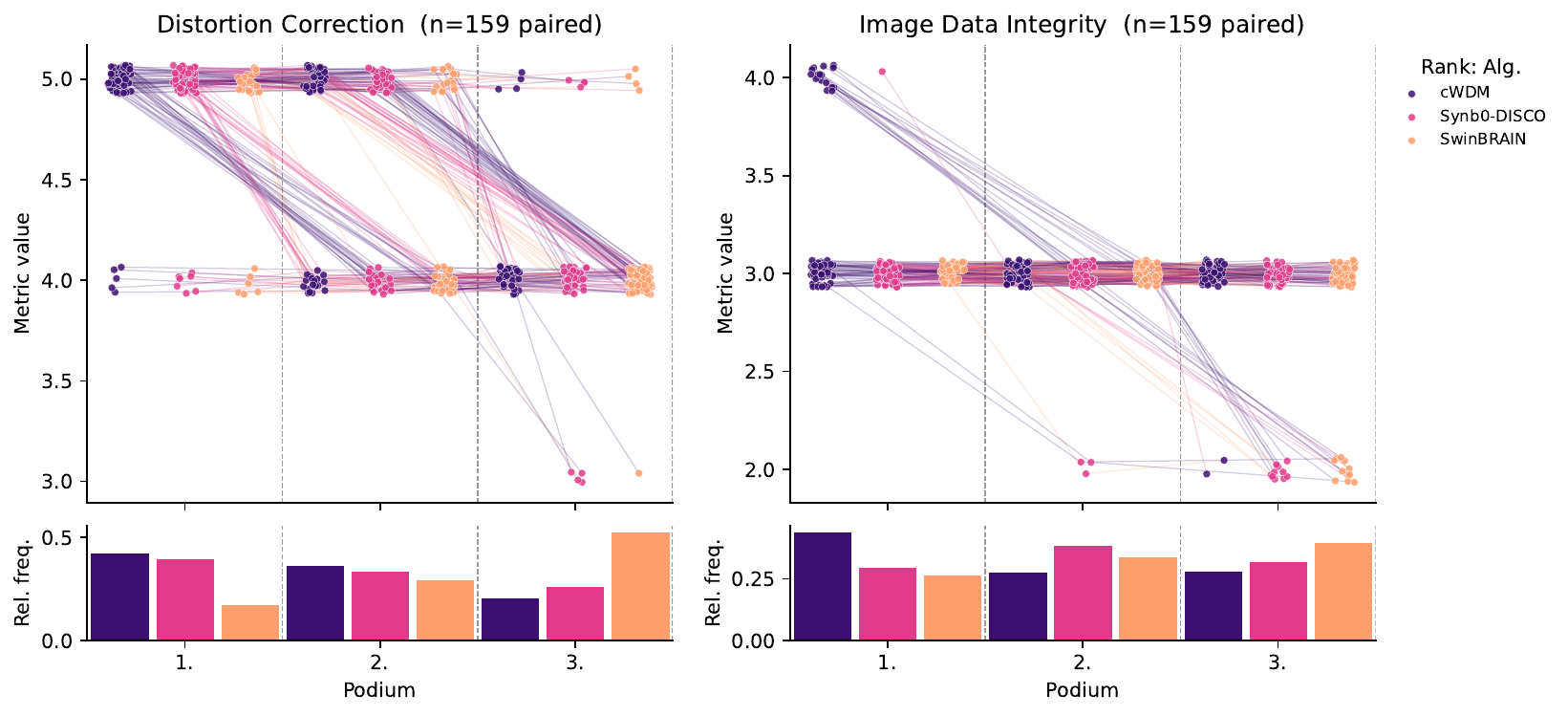}
        \vspace{0.5em}
        \textbf{(b)} Podium plots
        \label{fig:podium_plots}
    \end{minipage}

    \caption{Radiographer ratings and podium plots \cite{wiesenfarth2021methods} for the three models (cWDM, SwinBRAIN, Synb0-DISCO) across 159 test cases. (a) Distributions of ratings for Distortion Correction and Image Data Integrity. (b) Podium plots: top panel shows per-case metric values ranked 1st–3rd (best–worst), with lines connecting values from the same test case; bottom panel shows, for each algorithm, the relative frequency of attaining each rank.}
    \label{fig:qualitative_assessment_combined}
\end{figure}

\begin{figure}[!htb]
    \centering
    \includegraphics[width=0.70\textwidth]{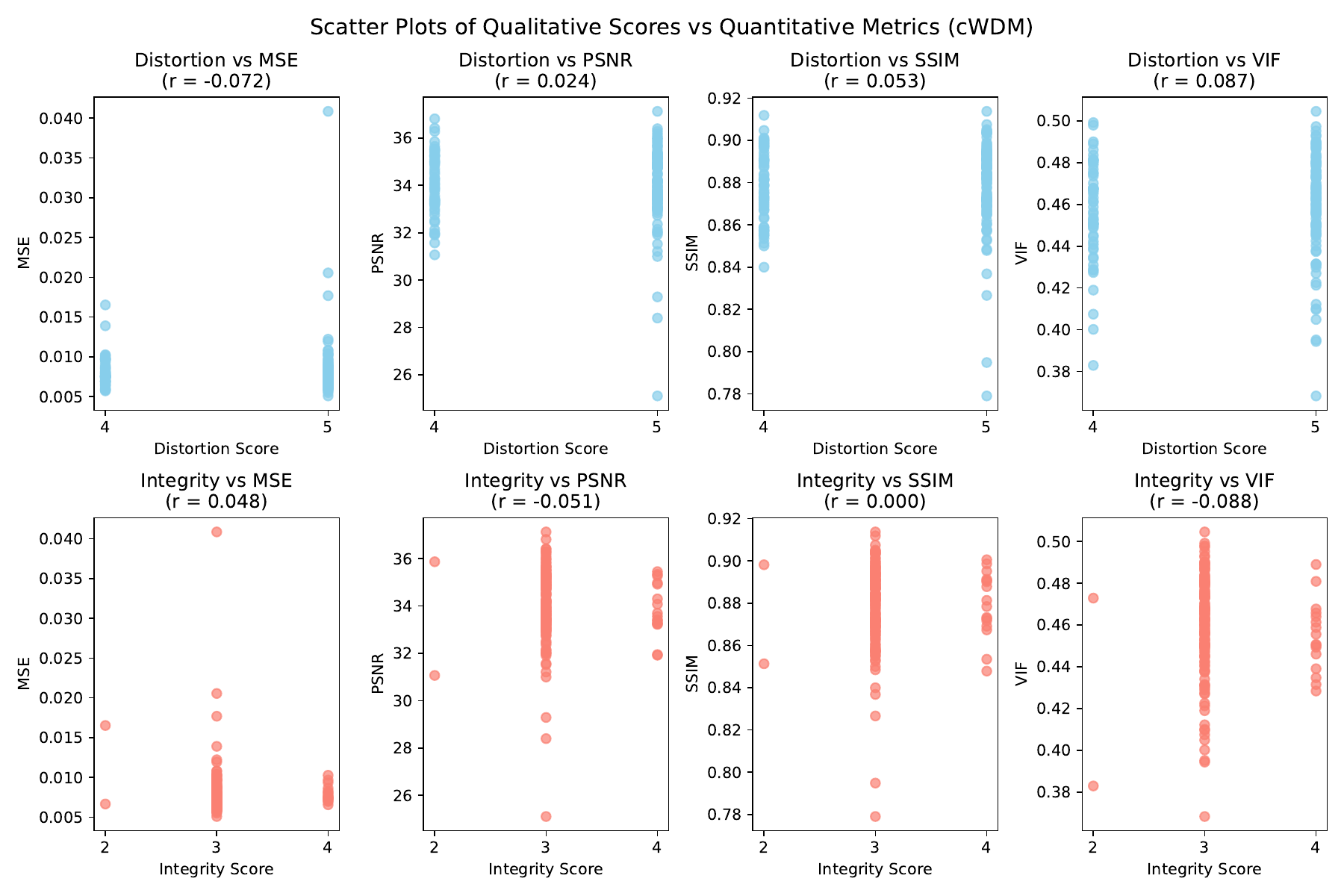}
    \caption{Scatter plots showing the relationship between qualitative evaluation scores (Distortion and Integrity) and quantitative image quality metrics (MSE, PSNR, SSIM, and VIF) for the cWDM model. Spearman correlation coefficients (r) are reported for each metric pair.}
    \label{fig:scatter_plots}
\end{figure}

Qualitative ratings (Fig.~\ref{fig:radiographer_ratings}) show cWDM outperforming the other models on both criteria: it achieved ``excellent'' distortion correction in 103/159 cases, versus 92 (Synb0-DISCO) and 45 (SwinBRAIN), and ``excellent'' image data integrity in 16 cases, versus 1 (SwinBRAIN) and 0 (Synb0-DISCO). Inter-rater agreement across three radiographers, assessed via Gwet's AC2 on 50 samples, was high for both Distortion Correction (AC2 = 0.835) and Image Data Integrity (AC2 = 0.945). Correlation analysis between the qualitative scores and quantitative metrics showed no statistically significant association as depicted in Figure~\ref{fig:scatter_plots}. This indicates that conventional image-similarity measures may not fully capture clinically relevant anatomical quality and may suggest the need for complementary qualitative evaluation. Based on observation of the sample results, the network appears to learn a combined mapping involving both distortion correction and spatial alignment. This is expected because the HCP corrected b0 targets are generated through preprocessing steps that include registration to the T1-weighted image and distortion correction.

\subsubsection{Results on LMIC Data}

\begin{figure}[H]
    \centering
    \includegraphics[width=0.75\textwidth]{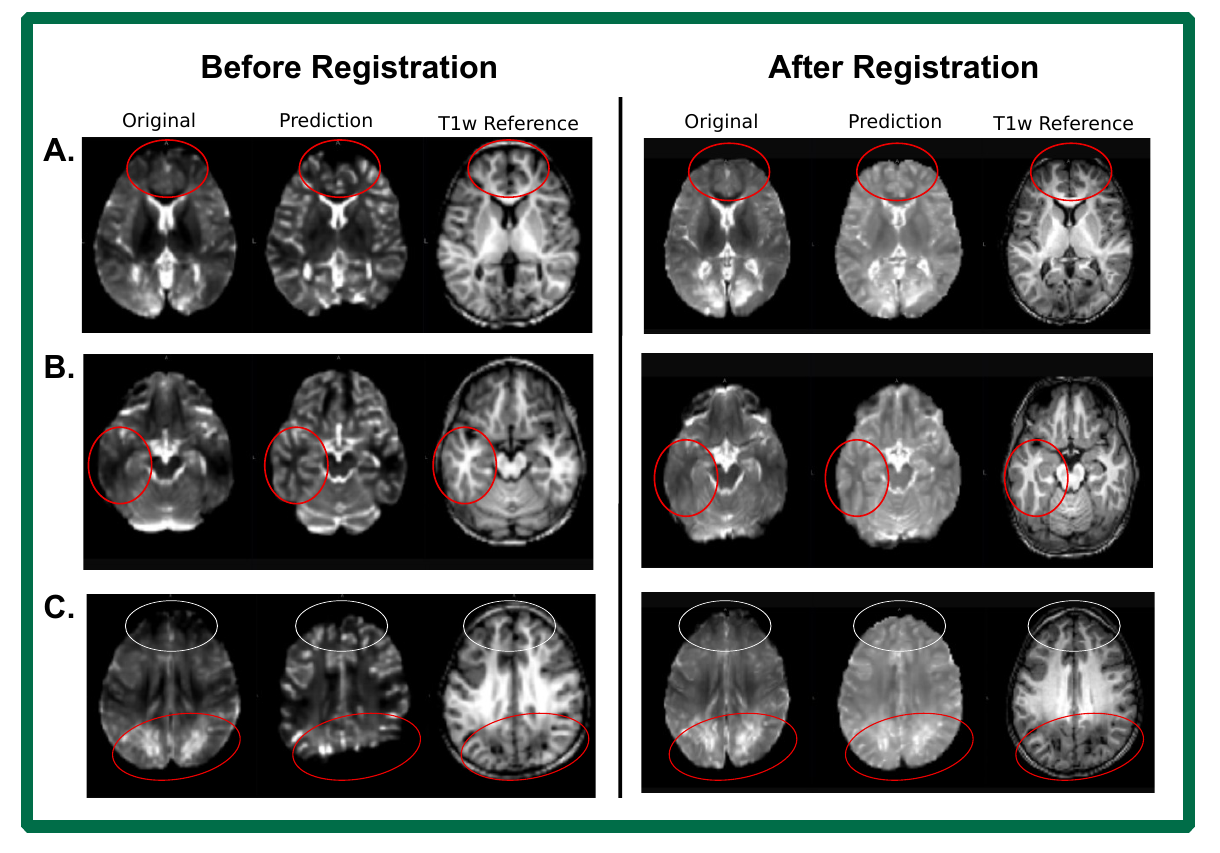}
    \caption{Qualitative comparison of predicted DWI before and after registration training. For three subjects (A–C), original DWI, model prediction, and T1w reference are shown pre- and post-registration. Red circles mark information loss (A: frontal, C: occipital) that resolves after registration, and reduced over-reliance on T1w structure, particularly temporal lobe (B). White circle (C) shows improved distortion correction accuracy post-registration. Contrast alteration persists across all images.}
    \label{fig:lmic_samples}
\end{figure}

After applying the best performing cWDM model on the LMIC data, acceptable distortion correction with loss of image information in some images and stronger mapping from the reference T1w image were observed, as per the qualitative assessment. Contrast alteration in overall image was also observed. Few samples are shown in Figure~\ref{fig:lmic_samples}. We hypothesized that this problem resulted from substantial domain shift. So, to minimize the population shift and few parameters of acquisition shifts like FOV and patient orientation, the model was retrained on the HCP dataset after linearly registering the distorted b0 image to the corresponding T1-weighted image, followed by nonlinear registration of both images to MNI152 space using EasyReg\cite{iglesias2023easyreg}, a deep learning tool. The same registration protocol was then applied to the LMIC dataset. This method improved the predictions in terms of both distortion correction and data integrity, as depicted in Figure~\ref{fig:lmic_samples}. No loss of image information was observed, however subtle mapping from T1w structure and contrast alteration was present. This suggests mapping both training and test data from different domains to a common standard space improves transferability in this task of distortion correction.

\textbf{Limitations}
Our evaluation is limited by two baselines (Synb0-DISCO and SwinBRAIN) alongside cWDM and does not include additional exhaustive benchmark contemporary learning-based distortion correction methods.
Additionally, our transferability analysis on the LMIC dataset is qualitative-only, due to unavailability of distortion-free reference acquisition. 
Our hypothesis on ``overreliance of T1-anatomy'' could have additionally been tested via T1-ablated or T1-mismatched retraining.

\subsubsection{Impact in Resource-Constrained Settings}
In resource-constrained settings, high patient throughput often forces limited per-patient scanner time resulting in reliance on rapidly-blipped single-shot EPI for routine DWI, trading image quality for speed.  
This causes geometric distortion from susceptibility differences at bone-air/tissue interfaces (e.g., skull base, orbits), causing pixel misregistration and warped anatomy. 
Established fixes - dual phase-encoding acquisition or vendor solutions like Siemens' RESOLVE - correct distortion but at the cost of longer scans and added licensing/hardware expense, making them impractical for high-volume, budget-constrained clinics. 
This leaves centers with a hard trade-off: fast but distorted DWI, or slow but corrected DWI. 
Our method targets this gap: by leveraging large-scale pretraining and generative deep learning for retrospective distortion correction, we push the state-of-the-art in clinically acceptable, anatomically faithful DWI without added scan time or vendor-specific MRI machines, offering a deployable solution for low-resource imaging workflows.

\section{Acknowledgement}
The authors would like to thank the following instructors of the Sprint AI Training for African Medical Imaging Knowledge Translation (SPARK) Academy 2026 summer school on deep learning in medical imaging for providing insightful background knowledge on brain tumors that informed the research presented here; The authors would also like to thank Linshan Liu for administrative assistance in supporting the SPARK Academy training and capacity-building activities, which the authors immensely benefited from. The authors acknowledge the computational infrastructure support from the Digital Research Alliance of Canada (The Alliance) and knowledge translation support from the McGill University Doctoral Internship program through the student exchange program for the SPARK Academy. In addition we thank NAAMII, Nepal for providing computational resources, office space, mentorship, and guidance in support of this project. Finally, we would like to thank Lacuna Fund for Health and Equity, the Radiological Society of North America (RSNA), Research \& Education (R\&E) Foundation Derek Harwood-Nash International Education Scholar Grant, and National Science and Engineering Research Council of Canada (NSERC) Discovery Launch Supplement (PI:UdunnaAnazodo, DGECR-2022-00136) for making the SPARK Academy possible via research grant supports.

\bibliographystyle{splncs04}
\bibliography{bibliography}

\end{document}